# Boosting The Accuracy of Multi-Spectral Image Pan-sharpening By Learning A Deep Residual Network

Yancong Wei, *Student Member, IEEE*, Qiangqiang Yuan, *Member, IEEE*， Huanfeng Shen, *Senior Member, IEEE*, Liangpei Zhang, *Senior Member, IEEE*

*Abstract*—In the field of multi-spectral and panchromatic images fusion (Pan-sharpening), the impressive effectiveness of deep neural networks has been recently employed to overcome the drawbacks of traditional linear models and boost the fusing accuracy. However, to the best of our knowledge, existing research works are mainly based on simple and flat networks with relatively shallow architecture, which severely limited their performances. In this paper, the concept of residual learning has been introduced to form a very deep convolutional neural network to make a full use of the high non-linearity of deep learning models. By both quantitative and visual assessments on a large number of high quality multi-spectral images from various sources, it has been supported that our proposed model is superior to all mainstream algorithms included in the comparison, and achieved the highest spatial-spectral unified accuracy.

## I. INTRODUCTION

Pan-sharpening is a fundamental and significant task in the field of remote sensed data fusion, in which high resolution spatial details from panchromatic (PAN) images and rich spectral information from multi-spectral (MS) or hyper-spectral (HS) images are fused to yield imagery with high resolution in both spatial and spectral domains. In this paper, our study focuses on the fusion of PAN and MS images.

Traditional pan-sharpening algorithms can be divided into three major branches: Component Substitution (CS)[1][2], Detail Injection (DI)[3][4], and regularization constraint model based methods[5][6]. In the former two branches, the fusing process is usually split into discrete steps, instead of end-to-end mapping. Thus, though such steps can be simply and quickly performed with impressively sharpened spatial details, obvious distortions are easily caused in the spectral domain, which severely degrade the spatial-spectral unified accuracy of images. Regularization constraint models describe the whole fusing process as linear functions of matrices or tensors with strict constraints, which are based on prior knowledge or reasonable assumptions about images included in the fusing process(such as total variaotion and sparse coding，etc), thus results with relatively higher quantitative accuracy can be produced. However, the performances of such models are limited by their linearity, which cannot accurately describe the fusing process that contains complex transformations in both domains. Besides, the reliance on prior constraints may also cause severe quality degeneration in cases where the prior knowledge does not fit the problem.

To overcome the drawbacks of previously proposed algorithms, inspired by the impressive performance of deep learning in the field of computer vision, we proposed a Deep Residual Pan-sharpening Neural Network (DRPNN) in this paper for robust and high quality fusion of PAN and MS images. The prototype of DRPNN is introduced from the field of image super-resolution[7], and we made a specific improvement on its architecture to fit to the task of image pan-sharpening. Supported by the residual learning architecture, an extremely deep convolutional filtering framework is formed to improve the accuracy of fusion, while the learning process of filtering parameters is also guaranteed to converge quickly.

The rest of this paper is organized as follows: Background knowledge about single image super-resolution, pan-sharpening and the superiority of deep learning is discussed in Section II, a detailed description on the proposed methodology is given in Section III, and results of experiments are listed and discussed in Section IV. After the conclusion in Section V, referenced research works are listed.

## II. BACKGROUND

**Single Image Super-Resolution (SISR)** aims to blindly predict a high resolution image $f$ from low resolution observation $g$. Due to the huge loss of information during the transformation from $f$ to $g$, the predicting process is highly ill-posed, and similarly to the previously discussed problem, the accuracy of solutions based on linear models are not as satisfying as those from non-linear models[7][8].

**Multi-Spectral Image Pan-sharpening:** We write a PAN image as $g_{PAN}$ (Size: $H \times W$) and a MS image with $S$ spectral bands as $g_{MS}$ (Size: $H/scale \times W/scale \times S$), the aim of pan-sharpening is yielding an image with high resolution in both domains, and we write it as $f_{MS}$ ($H \times W \times S$). The fusing process is regarded as guided super-resolution, while in the spatial domain of $g_{MS}$, the illness of the inverse predicting problem is reduced with $g_{PAN}$ included, compared to the blind situation in single image super-resolution.

The main difficulty of pan-sharpening comes from the spectral domain, as the bandwidths covered by PAN and MS channels are not guaranteed to fully overlap among various types of sensor, e.g. WorldView-2 (PAN: 400nm-1040nm, MS:

Continuously covering 450nm-800nm) and IKONOS (PAN: 526nm-929nm, MS: Discretely covering 445nm-853nm). Thus, to preserve the spectral fidelity of images fused from such observations, the merging process in the spectral domain is also very complex and needs to be simulated using highly non-linear functions.

**Pan-sharpening Based On Deep Learning:** To break the limit of linear models discussed above, deep neural networks are recently employed in many advanced works to perform end-to-end prediction and yield images with state-of-the-art accuracy, which relies on the non-linearity of the mapping process through deep networks. Similarly to the successful applications to SISR[7][8], the branch of deep learning models has also been introduced to the field of image fusion by some previously proposed works[9][10].

By using bicubic interpolation to coarsely up-sampling $g_{MS}$ to $G_{MS}$ ( $H \times W \times S$ ), we have an initialized input $G = \{G_{MS}, g_{PAN}\}$ ( $H \times W \times (S+1)$ ) for the pan-sharpening task. The high resolution MS image can be reconstructed by extracting various features from related components of $G$: Low frequency features from $G_{MS}$ and high frequency features from $g_{PAN}$, then merging them to form the final estimation.

Thus, from the perspective of deep learning, we conclude the whole process as a filtering function with high non-linearity, and the requirements can be met well by the nature of deep neural networks, in which multiple linear filtering layers are stacked to form a highly non-linear transformation, and the optimal allocations for all parameters can be automatically searched to minimize the predicting loss between output of the network $F = T(G)$ and the ground truth $f_{MS}$. The whole flowchart of learning a deep network to the pan-sharpening process has been described in Fig. 1.

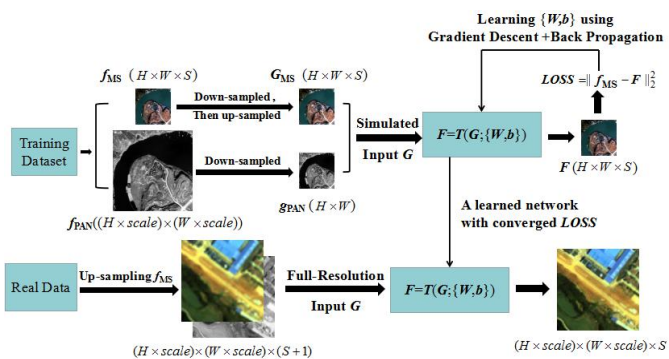

Fig. 2. The framework of pan-sharpening based on deep learning.

## III. METHODOLOGY

**Convolutional Neural Network (CNN):** CNN is one of the most impressive branch of deep learning models in the field of computer vision. In an end-to-end image restoration task, a CNN built with multiple stacked convolutional layers functions as $f \approx F = CNN(G)$ to estimate the high quality image $f$ from the degraded observation $G$. To train a randomly initialized CNN for pan-sharpening, down-sampled MS and PAN images are inputted as $G$, then an image $F$ with the same size as original MS image $f_{MS}$ is produced[10]. In a CNN with $L$ layers, the fusing process is performed via forward-passing:

$$F_0 = G, F_l = \max(0, W_l \circ F_{l-1} + b_l), \quad l = 1,...,L-1 \quad (1)$$

$$F = W_L \circ F_{L-1} + b_L \quad (2)$$

where $W$ stands for three-dimensional convolutional filters and $b$ represents bias vectors. With Stochastic Gradient Descent (SGD) and Back Propagation (BP), all parameters $\{W, b\}$ in a network can be iteratively learned to reach an optimal allocation. The learning process goes as:

$$\delta\theta^t = \{\delta W^t, \delta b^t\} = \{\frac{\partial(\|f_{MS} - F^t\|_2^2)}{\partial W^t}, \frac{\partial(\|f_{MS} - F^t\|_2^2)}{\partial b^t}\} \quad (3)$$

$$\theta^{t+1} = \theta^t + \Delta\theta^t = \theta^t + \mu \cdot \Delta\theta^{t-1} - \varepsilon \cdot \delta\theta^t \quad (4)$$

**Deep Residual Learning:** It has been indicated that a deeper CNN with more filtering layers tends to extract more abstract and representative features[11][12], thus higher predicting accuracy can be expected. But due to the gradient vanishing problem, the gradients of predicting loss to parameters in shallow layers can not be smoothly passed via BP[7][8][10], which disturbs a deep network from being fully learned.

Deep residual learning[13] is an advanced method for this problem, in which the transformation $f \approx CNN(G)$ is replaced with $f - G \approx RES(G)$ by setting a skip connection between disconnected layers. It is reasonable to assume that most pixel values in the residual image $f - G$ are very close to zero and the spatial distribution of residual features should be very sparse, which can transfer the gradient descending process to a much smoother hyper-surface of loss to filtering parameters. Thus, searching an allocation that is very close to the optimal for $\{W, b\}$ becomes much faster and easier, which allows us to add more layers to the network and boost its performance.

However, in pan-sharpening tasks, it should be noted that the size of the final output $F$ ( $H \times W \times S$ ) is not the same as the size of the input $G$ ( $H \times W \times (S+1)$ ), thus, instead of directly predicting the end-to-end residual image[7][14], the process through DRPNN with $L$ layers is divided into two stages:

**Stage 1**: The 1st to $(L-1)$-th layers are stacked under a skip-connection to estimate the residual between $G$ and $F^{Stage\ 1}$ (Size: $H \times W \times (S+1)$ ). The convolutional filtering process in each layer is the same as described as (1), then the residual outputted from the $(L-1)$-th layer is added to $G$ to yield $F^{Stage\ 1}$:

$$F^{Stage\ 1} = G + F_{L-1} \quad (5)$$

**Stage 2**: The $L$-th layer of DRPNN is set to reduce the spectral dimensionality from ($S+1$) bands to $S$ bands via the last time of convolutional filtering in the network, after which the final output $F = F^{Stage\ 2}$ ( $H \times W \times S$ ) is yielded:

$$F = F^{Stage\ 2} = W_L \circ F^{Stage\ 1} + b_L \quad (3)$$

The complete architecture of DRPNN has been illustrated in Fig. 2.

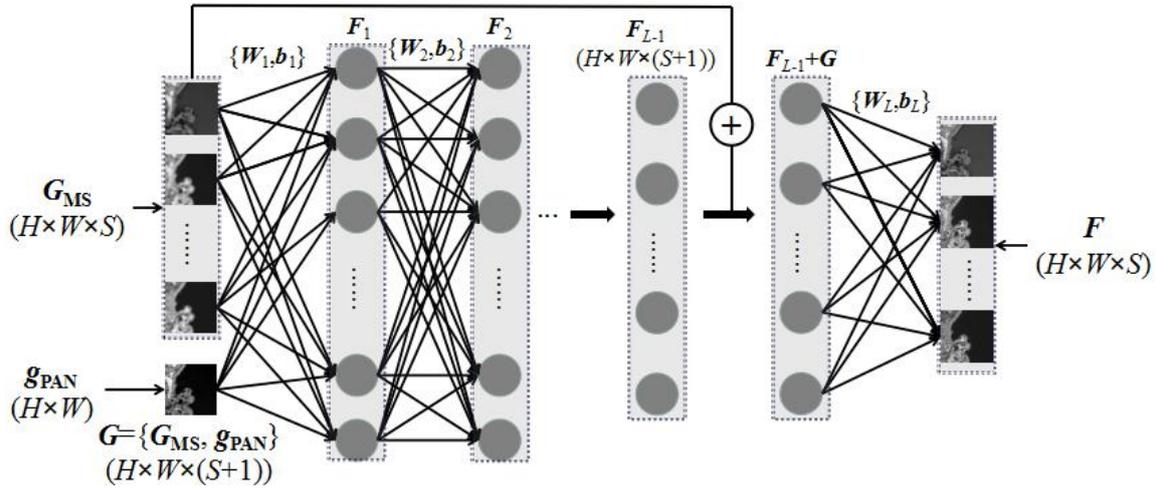

Fig. 2. The whole flowchart of passing low resolution MS and PAN images through a Deep Residual Pan-sharpening Neural Network.

## IV. EXPERIMENTS AND DISCUSSIONS

**Datasets:** For simulated training and testing of DRPNN, two datasets are collected from images obtained by QuickBird and WorldView-2. Two networks are separately built and learned for MS images with different $S$: One is set with $S = 4$ (QuickBird), the other is set with $S = 8$ (WorldView-2). Besides, for real-data experiments, we also collected a dataset from a group of high quality IKONOS images to perform real-data experiments for the network with $S = 4$, while the other model is still tested on the WorldView-2 dataset, but fed with full-resolution MS images.

**Hyper-Parameters:** A DRPNN proposed in this paper contains $L = 11$ layers, the $l$-th of which is filled with $C_l$ groups of filters $W_{l,k}(h_l \times w_l \times C_{l-1})$, where $k = 1,...,C_l$, and one bias vector $b_l(1 \times C_l)$. According to the aimed task, $C_0 = S+1$ and $C_{11} = S$ should be provided at first, and similarly to the prototype[7], $C_l$ for the rest layers is empirically set to 64. For the spatial size $h_l \times w_l$ of filters, we have compared the performances of several values and showed them in Fig. 3. Indicated by the results of comparison, the filter size is set to $7 \times 7$ through the whole network,

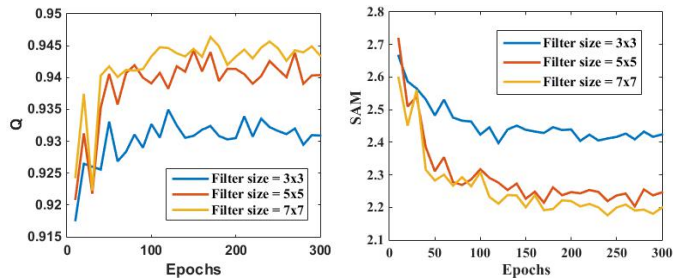

Fig. 3. Q and SAM of DRPNN with different filter sizes, the quantitative assessments are simulated on the dataset of 160 QuickBird images.

**Training:** The training process of each network cost 300 epochs with a batchsize of 64 for applying SGD. The Classic Momentum (CM) algorithm[15] is also applied to accelerate the descending of the prediction loss, where the learning rate $\varepsilon$ is initialized to 0.05 for the first 10 layers under the skip connection and 0.005 for the last layer, while the momentum $\mu$ is fixed to 0.95 for the whole network. By every 60 epochs, the learning rate is multiplied by a descending factor $\gamma = 0.5$. The implementation of convolutional neural networks is supported by two deep learning frameworks: Caffe[16] for training and MatConvNet[17] for testing.

**Quantitative Assessments:** MS and PAN images are down-sampled to simulate the relatively degenerated input $G$. Four metrics: Q, ERGAS, SAM, and SCC are employed to quantify its accuracy in spatial and spectral domains, with the original MS image as ground truth. The performances of DRPNN are compared with five algorithms from different branches for comparisons with the proposed network, including Component Substitution: GS[1], Detail Injection: MTF-GLP[3], SFIM[4], regularization constraint models: ISTS[5] based on total variation and TSSC[6] based on sparse representation. Besides these traditional algorithms, PNN[10], a shallow CNN without recidual learning and skip connection has also been included for comparison. Quantitative results are listed in Table I.

**TABLE I. QUANTITATIVE RESULTS OF SIMULATED EXPERIMENTS.**

| Tested images | Method | Q (↑) | ERGAS (↓) | SAM (↓) | SCC (↑) |
|---|---|---|---|---|---|
| Sensor: QuickBird Size: 250×250×4 Total number: 160 | GS | 0.8305 | 4.5014 | 4.0227 | 0.6090 |
| | MTF-GLP | 0.8227 | 4.4409 | 3.7893 | 0.5839 |
| | SFIM | 0.8264 | 5.1491 | 3.7708 | 0.5708 |
| | ISTS | 0.8498 | 3.8677 | 3.6579 | 0.6157 |
| | TSSC | 0.8488 | 3.9773 | 3.7154 | 0.5909 |
| | PNN | 0.9206 | 2.7110 | 2.6405 | 0.7951 |
| | DRPNN | **0.9437** | **2.1916** | **2.1936** | **0.8458** |
| Sensor: WorldView-2 Size: 250×250×8 Total number: 80 | GS | 0.8606 | 4.8395 | 6.1412 | 0.8190 |
| | MTF-GLP | 0.8788 | 4.3748 | 5.7698 | 0.8136 |
| | SFIM | 0.8756 | 4.3230 | 5.7579 | 0.8281 |
| | ISTS | 0.8720 | 4.4353 | 5.8933 | 0.8126 |
| | TSSC | 0.8951 | 3.9735 | 5.8269 | 0.8116 |
| | PNN | 0.9389 | 3.0695 | 4.4757 | 0.8674 |
| | DRPNN | **0.9458** | **2.8913** | **4.1998** | **0.8766** |

The comparisons of numeric metrics have told that the DRPNN yielded images with the best spatial-spectral unified accuracy. However, while the numeric metrics are mainly employed to quantify the overall accuracy of a predicted image, visual inspections are also required to find noticeable distortions, which may elude from quantitative assessments.

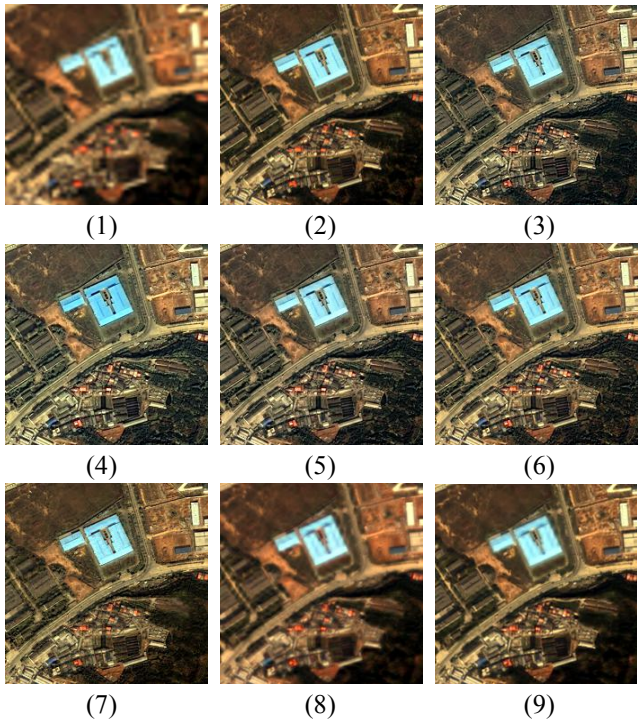

Fig. 4. Simulated fusion results from QucikBird, Yichang, 2015. (1) Low resolution MS simulated by down-sampling; (2) Ground truth; (3) SFIM; (4) GS; (5) MTF-GLP; (6) ISTS; (7) TSSC; (8) PNN; (9) Ours.

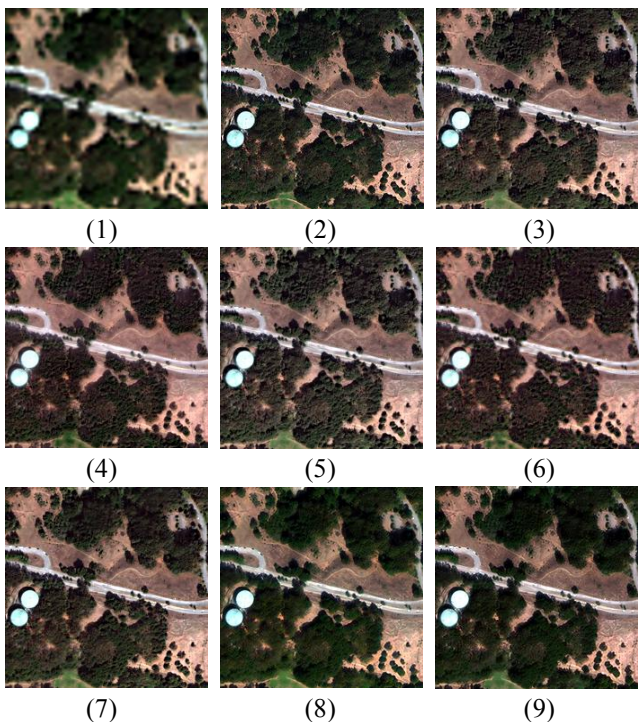

Fig. 5. Simulated fusion results from WorldView-2, San Francisco, 2011. (1) Low resolution MS simulated by down-sampling; (2) Ground truth; (3) SFIM; (4) GS; (5) MTF-GLP; (6) ISTS; (7) TSSC; (8) PNN; (9) Ours.

**Visual Assessments:** From each dataset, one group of simulated fusion results are picked out and displayed as true-color images in Fig. 4 and Fig. 5. Comparing the results from PNN and DRPNN with those from other algorithms, it can be observed that the extremely sharpened spatial features in Fig.5 (3)-(7) are achieved with severe spectral distortions, while the two deep learning based models performed the task with highest similarities to the ground truths, both on the merging of spatial details and the preservation of spectral fidelity, even for some specific spectral curves that can be easily distorted, e.g. The bare soils in Fig. 5 (8)-(9).

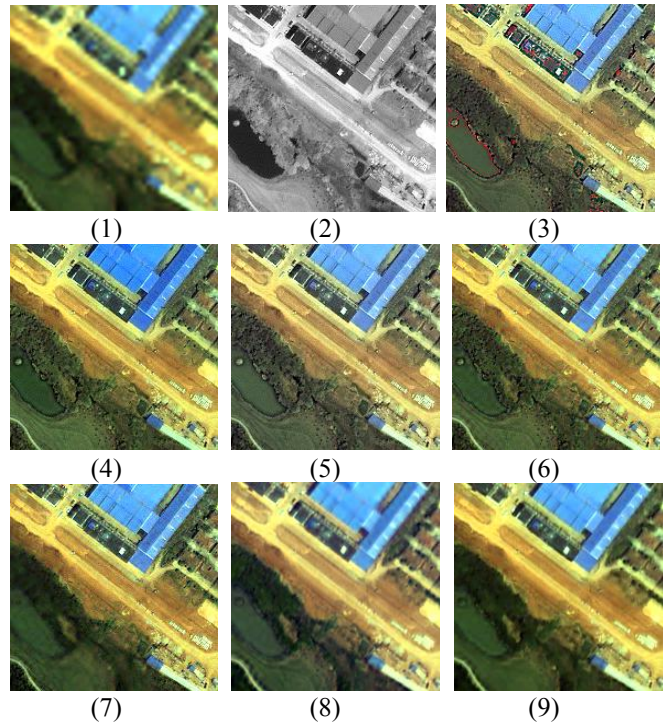

Fig. 6. Full-resolution fusion results from IKONOS, Wuhan. (1) Up-sampled MS; (2) PAN; (3) SFIM; (4) GS; (5) MTF-GLP; (6) ISTS; (7) TSSC; (8) PNN; (9) Ours.

For the two CNNs, the high quality results from them are slightly difficult to be distinguished, but by investigating the preservation of ground objects with small sizes, it can be confirmed that the deeper architecture of DRPNN has made certain contributions to more appropriately sharpening the small edges, e.g. The edge of industrial areas in the lower left part of the scene in Fig. 4 (8)-(9). The same tendency has also been indicated in the full-resolution results of real-data experiments, a group of which has been displayed in Fig. 6.

## V. CONCLUSION

In this paper, we proposed the Deep Residual Pansharpening Neural Network to performance high quality fusion of MS and PAN images. Compared to traditional linear model based methods, the DRPNN employs the high non-linearity of CNN to achieve much higher accuracy, and to make adequate use of the advantages of deep learning, the residual learning architecture is applied to allow the network going deeper and boost its performance. The superiority of the proposed network is supported by results of experiments on a large number of images covering various and complex ground scenes.

The successful implementation of the DRPNN in this paper motivated us to apply the framework to further studies in the field of multi-source remote sensed data fusion. For instance, the accuracy of MS and HS image fusion can also be improved by using three-dimensional CNN[18], in which the extremely high dimensionality of hyper-spectral images are simply reduced by principle component analysis (PCA), while a recent study[19] has indicated that saliency-based band selection can be studied to compress and better represent the rich spectral information. Thus, as our future works tend to process hyper-spectral data (Quality restoration, fusion, interpretation), the combination of feature learning based on neurocomputing and saliency detection based on manifold feature representation[20][21] shall be focused in particular.

REFERENCES

[1] C. A. Laben and B. V. Brower, "Process for enhancing the spatial resolution of multispectral imagery using pan-sharpening," ed: Google Patents, 2000.

[2] J. Choi, K. Yu and Y. Kim, "A New Adaptive Component-Substitution-Based Satellite Image Fusion by Using Partial Replacement," in *IEEE Transactions on Geoscience and Remote Sensing*, vol. 49, no. 1, pp. 295-309, Jan. 2011.

[3] B. Aiazzi, L. Alparone, S. Baronti, A. Garzelli, and M. Selva, "MTF-tailored multiscale fusion of high-resolution MS and pan imagery," *Photogrammetric Engineering and Remote Sensing*, vol. 72, pp. 591-596, May 2006.

[4] SFIMJ. G. Liu, "Smoothing Filter-based Intensity Modulation: a spectral preserve image fusion technique for improving spatial details," *International Journal of Remote Sensing*, vol. 21, pp. 3461-3472, Dec 2000.

[5] H. Shen, X. Meng and L. Zhang, "An Integrated Framework for the Spatio‐Temporal‐Spectral Fusion of Remote Sensing Images," in *IEEE Transactions on Geoscience and Remote Sensing*, vol. 54, no. 12, pp. 7135-7148, Dec. 2016.

[6] C. Jiang, H. Zhang, H. Shen and L. Zhang, "Two-Step Sparse Coding for the Pan-Sharpening of Remote Sensing Images," in *IEEE Journal of Selected Topics in Applied Earth Observations and Remote Sensing*, vol. 7, no. 5, pp. 1792-1805, May 2014.

[7] J. Kim, J. K. Lee and K. M. Lee, "Accurate Image Super-Resolution Using Very Deep Convolutional Networks," *2016 IEEE Conference on Computer Vision and Pattern Recognition (CVPR)*, Las Vegas, NV, 2016, pp. 1646-1654.

[8] C. Dong, C. C. Loy, K. He and X. Tang, "Image Super-Resolution Using Deep Convolutional Networks," in *IEEE Transactions on Pattern Analysis and Machine Intelligence*, vol. 38, no. 2, pp. 295-307, Feb. 1 2016.

[9] W. Huang, L. Xiao, Z. Wei, H. Liu and S. Tang, "A New Pan-Sharpening Method With Deep Neural Networks," in *IEEE Geoscience and Remote Sensing Letters*, vol. 12, no. 5, pp. 1037-1041, May 2015.

[10] G. Masi, D. Cozzolino, L. Verdoliva, and G. Scarpa, "Pansharpening by Convolutional Neural Networks," *Remote Sensing,* vol. 8, Jul 2016.

[11] D. Erhan, Y. Bengio, A. Courville, and P. Vincent, "Visualizing higher-layer features of a deep network," *University of Montreal,* vol. 1341, p. 3, 2009.

[12] J. Yosinski, J. Clune, A. Nguyen, T. Fuchs, and H. Lipson, "Understanding neural networks through deep visualization," *arXiv preprint arXiv:1506.06579,* 2015.

[13] K. He, X. Zhang, S. Ren and J. Sun, "Deep Residual Learning for Image Recognition," *2016 IEEE Conference on Computer Vision and Pattern Recognition (CVPR)*, Las Vegas, NV, 2016, pp. 770-778.

[14] K. Zhang; W. Zuo; Y. Chen; D. Meng; L. Zhang, "Beyond a Gaussian Denoiser: Residual Learning of Deep CNN for Image Denoising," in *IEEE Transactions on Image Processing* , vol.PP, no.99, pp.1-1.

[15] B. T. Polyak, "Some methods of speeding up the convergence of iteration methods," *Ussr Computational Mathematics & Mathematical Physics,* vol. 4, pp. 1-17, 1964.

[16] Jia, E. Shelhamer, J. Donahue, S. Karayev, J. Long, R. Girshick*, et al.*, "Caffe: Convolutional architecture for fast feature embedding," in *Proceedings of the 22nd ACM international conference on Multimedia*, 2014, pp. 675-678.

[17] MatConvNet: CNNs for MATLAB. Available online: http://www.vlfeat.org/matconvnet.

[18] F. Palsson, J. R. Sveinsson and M. O. Ulfarsson, "Multispectral and Hyperspectral Image Fusion Using a 3-D-Convolutional Neural Network," in *IEEE Geoscience and Remote Sensing Letters*, vol. 14, no. 5, pp. 639-643, May 2017.

[19] Q. Wang, J. Lin and Y. Yuan, "Salient Band Selection for Hyperspectral Image Classification via Manifold Ranking," in *IEEE Transactions on Neural Networks and Learning Systems*, vol. 27, no. 6, pp. 1279-1289, June 2016.

[20] Q. Wang, Y. Yuan and P. Yan, "Visual Saliency by Selective Contrast," in *IEEE Transactions on Circuits and Systems for Video Technology*, vol. 23, no. 7, pp. 1150-1155, July 2013.

[21] Q. Wang, Y. Yuan, P. Yan and X. Li, "Saliency Detection by Multiple-Instance Learning," in *IEEE Transactions on Cybernetics*, vol. 43, no. 2, pp. 660-672, April 2013.